\documentclass{ifacconf}

\usepackage{natbib}        % required for bibliography

\usepackage{cite}
\usepackage{amsmath,amssymb,amsfonts}
\usepackage{algorithmic}
\usepackage{graphicx} 
\usepackage{color}
\usepackage{xcolor}
\usepackage{subcaption}
\usepackage{float}
\usepackage{diagbox}
\usepackage{comment}

\begin{document}

\begin{frontmatter}
\title{One For All: LLM-based Heterogeneous   Mission Planning in Precision Agriculture}
% Below are some alternative titles suggested by AI
% - One For All: Unified LLM-Driven Mission Planning for Heterogeneous Agricultural Robotics
% - One For All: Large Language Models Orchestrating Multi-Robot Precision Agriculture Missions
% - One For All: A Generative AI Approach to Coordinated Robotic Task Planning in Agriculture
% - One For All: Transforming Agricultural Robotics with Large Language Model-Enabled Mission Coordination
% - One For All: Adaptive Mission Planning for Heterogeneous Robotic Systems in Precision Agriculture

\thanks[footnoteinfo]{
This matrial is based upon work supported  by the National Science Foundation (NSF) under Cooperative Agreement Number EEC-1941529
(IoT4Ag) and award CMMI-2326310, and by USDA-NIFA under award 
\#2021-67022-33452. Any opinions, findings, conclusions, or recommendations expressed in this publication are those of the author(s) and do not necessarily reflect the views of the NSF and USDA.}

\author[First]{Marcos Abel Zuzu\'{a}rregui}
\author[Second]{Mustafa Melih Toslak}
\author[First]{Stefano Carpin}

\address[First]{Department of Computer Science and Engineering\\ University of California, Merced, CA, USA.}
\address[Second]{Department of Informatics, Bioengineering, Robotics and Systems Engineering, University of Genova, Italy}

\begin{abstract} 
% original version

%The use of artificial intelligence in precision agriculture stands as an attempt to reinvent the way farmers can handle their daily tasks.
%Every day there is a new tool that simplifies the user experience. 
%However, with each new utility, we introduce a new learning curve.
%Often for non-technical users, one new tool to learn paired with their daily work is already too much.
%With our latest creation of a natural language (NL) robotic mission planner, we present a paper that demonstrates the flexibility of our architecture to support multiple robots through a single interface: no user configuration required.
%With the system described in this paper, non specialists 
%can formulate complex missions related to precision agriculture 
%without having to write and line of code, because by leveraging 
%large language models and a set of predefined primitives our
%architecture seamlessly translates human language into intermediate
%descriptions that can be ingested and executed by different robots.
%Through a new class of experiments using a manipulator to demonstrate %computer vision and manipulation mission plans, we show that our previous mission planner for wheeled robots is general enough to support an entire fleet of diverse autonomous robots, yet powerful enough to support complex mission requests. 

% claude's version + SC's version  % THIS COULD BE FINAL AS IS
Artificial intelligence is transforming precision agriculture, offering farmers new tools to streamline their daily operations. 
While these technological advances promise increased efficiency, they often introduce additional complexity and steep learning 
curves  that are particularly challenging for non-technical users who must balance tech adoption with existing workloads.
In this paper, we present a natural language (NL) robotic mission planner that enables non-specialists to control heterogeneous robots through a common interface. 
By leveraging large language models (LLMs) and predefined primitives, our architecture seamlessly translates human language into intermediate descriptions that can be executed by different robotic platforms. 
With this system, users can formulate complex agricultural missions without writing any code.
In the work presented in this paper,  we extend our previous system tailored for wheeled robot mission planning through a new class of experiments involving robotic manipulation and computer vision tasks. 
Our results demonstrate that the architecture is both general enough to support a diverse set of robots and powerful enough to execute complex mission requests. 
This work represents a significant step toward making robotic automation in precision agriculture more accessible to non-technical users.
\end{abstract}


\end{frontmatter}

\section{Introduction} % THIS COULD STAND AS IS

\begin{figure}[t]
\centering
\includegraphics[width=0.8\linewidth]{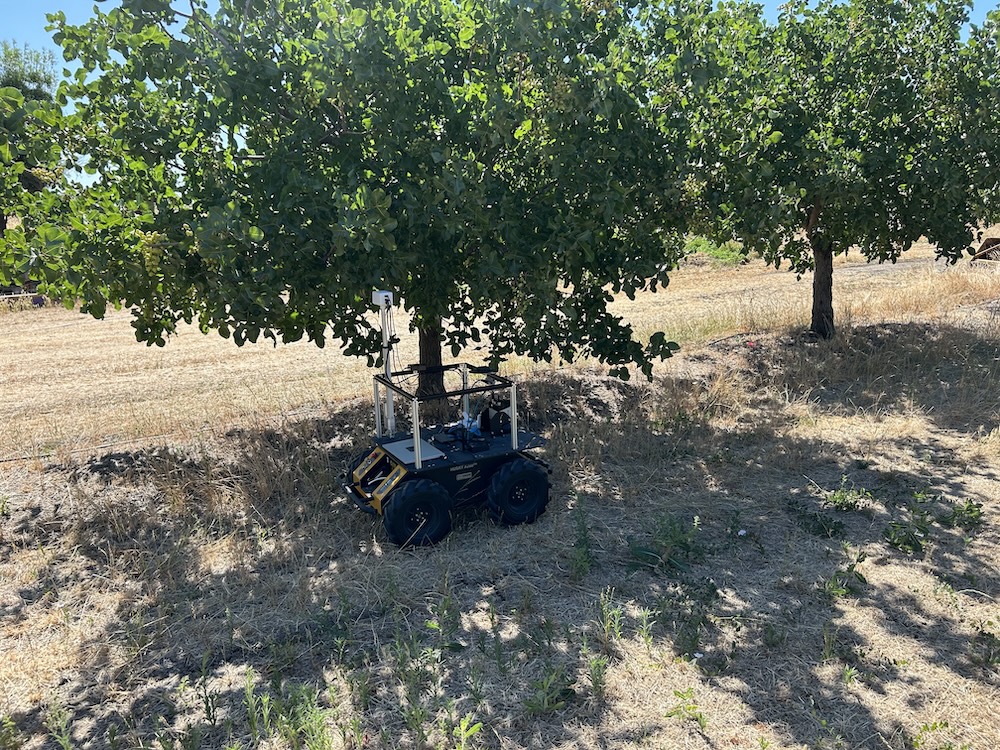}
\caption{Robot deployed in a pistachio orchard for water stress monitoring.}
\label{fig:graph20_ucm}
\end{figure}

Robotic mission planning (MP) represents an open challenge in robotics and autonomy, serving as a critical interface between user intent and desired outcomes. 
In precision agriculture, users are often non-technical specialists with specific goals, such as collecting soil samples or visually mapping an orchard.
Perhaps unsurprisingly, creating feasible robotic mission plans is far more complex than it might initially appear. 
The process involves significant challenges in decomposing mission goals into intermediate tasks and developing an interface that balances computational power with user-friendly design. 
While recent research has provided sophisticated solutions (see Section \ref{sec:sota} for more details), our ongoing work in this domain emphasizes generalization and standardization of mission planning approaches.
We recently introduced a large language model (LLM)-powered robot mission planner \citep{CarpinLLMSubmitted} specifically designed for farm orchards, initially testing it with a wheeled robot equipped with proximal sensing capabilities to detect water stress (Figure \ref{fig:graph20_ucm}). 
A key innovation is the ability to control the robot through simple natural language prompts.
This approach addresses critical challenges in precision agriculture, particularly in scenarios with limited network connectivity. 
These constraints require robots to execute missions autonomously, without the ability to periodically reconnect to cloud services, refine plans dynamically, or acquire additional information to deal with unexpected events.
This is in stark contrast to similar systems proposed for industrial or domestic robotic applications, where
connectivity is a given.
In this paper we expand our software architecture to formulate MP for heterogeneous robotic platforms. 
We are particularly interested in generating comprehensive orchard blueprints, also known as digital twins \citep{PURCELL2023100094},  with bidirectional data flow.
Methods for digital twin generation include remote geobiometric sensing and computer vision reconstruction via point clouds.
However, significant technical challenges remain. 
Because connectivity is constrained to specific points within our testbed,
we developed a one-shot planning approach without the assumption of feedback flowing back to the end user
while the mission unfolds.
This operational limitation drives our continued exploration of robust, adaptable robotic mission planning strategies for precision agriculture.\\
Developing an effective digital twin for agricultural applications requires integrating diverse sources of input data.
Consequently, smart farms operators must often engage with various robotic systems to formulate mission plans and acquire essential data on farm health and operational status.
Despite advances in precision agriculture, many emerging applications feature disparate interfaces tailored to specific robotic systems, limiting cross-compatibility.
To address these challenges, we have implemented a mission planning pipeline using the same architectural framework as \citep{CarpinLLMSubmitted}, now adapted for deployment with a Kinova KORTEX Gen3 manipulator.
Building upon our prior work focused on in-orchard sensing with wheeled robotic platforms using a Clearpath's Husky robot \citep{CARPINICRA2024D}, this study now investigates the feasibility of NL-based mission planning with a computer vision-enabled manipulation robot.
The overarching objective is to streamline robotic operations beyond controlled laboratory environments by enabling unified system control for both mobile and manipulator-based robotic platforms.
As we have previously worked to understand questions about a LLM being capable in an unstructured environment, we now ask whether or not this architecture is general enough for use on robots with different
capabilities while being powerful enough for complex mission queries.
In this paper, we extend the use of \citep{CarpinLLMSubmitted} to generate mission plans, using NL, for manipulators.
The contributions of this paper are the following:
\begin{itemize}
\item we present an LLM to robotic task execution pipeline for manipulator and computer-vision mission generation;
\item we investigate whether this architecture is flexible enough to support both wheeled robots and manipulators;
\item we validate our proposed system in the field and show its limits and strengths.
\end{itemize}
The rest of the paper is as follows.
Selected related work is presented in Section \ref{sec:sota}.
In Section \ref{sec:method} where we describe the system we developed, experiments detailing our findings are given in Section \ref{sec:results}, and conclusions are given in Section \ref{sec:conclusions}.
\section{Related Literature} \label{sec:sota}

In recent years, there has been a major increase in the number of applications leveraging LLMs to simplify the interface between user and execution. 
In computer science, the focus is often on NL mission plans to code generation \citep{ahn_as_2022, huang_language_2022, mower_ros-llm_2024, kannan_smart-llm_2024}.
The theme of these papers is to rely on semantic mapping of the LLM and use a mission query to plan and execute tasks. 
While the approaches of \citep{mower_ros-llm_2024, kannan_smart-llm_2024, liang_code_2023, huang_language_2022} leverage the generative capabilities of LLMs to handle complex mission planning and execution, they often encounter limitations due to inherent inaccuracies in language model outputs. 
These papers have shown, along with numerous papers on LLM drawbacks \citep{emsley_chatgpt_2023, ray_chatgpt_2023, kambhampati_llms_2024}, that planning and executing using an LLM is extremely difficult.
In contrast, our approach employs LLMs solely to generating initial task sequences within mission plans. 
To mitigate potential inaccuracies and ensure robust task definitions, we impose constraints on the LLM-generated outputs by validating them against predefined Extensible Markup Language (XML) schema definitions (XSD). 
Critically, none of the works cited in LLM MP take into consideration planning scenarios with limited network connectivity.
The solutions demonstrate the ability of task execution to reprompt the LLM for an updated plan.
Our work focuses on one-shot planning: planning that occurs only upon the initial mission plan generation.

The notion of accounting for LLM output uncertainty continues to be studied.
\citep{jousselme_uncertain_2023, pelucchi_chatgpt_2023} attempt to mitigate uncertainty through prompting frameworks, constraining the answers that ChatGPT can give based on available information.
While this paper does not fully dive into uncertainty computation, it does acknowledge LLMs often unstructured and incorrect outputs.
Previously, robot planning was done using languages such as linear temporal logic (LTL) \citep{jansen_can_2023} or Planning Domain Definition Language (PDDL) \citep{noauthor_ieee_2024}.
Instead, our architecture relies on XSDs to constrain the LLM output when generating a mission plan based on NL.
To the best of our knowledge, this is the first architecture that uses any software-based validation engine to ensure that the response of the mission planner fits a known syntax.

In \citep{mower_ros-llm_2024, kannan_smart-llm_2024}, each provide their own mechanism for task decomposition and system connectivity. 
Instead, we follow \citep{noauthor_ieee_2024} as our system framework paired with a simple implementation of a behavior tree.
This creates a generic and modular architecture that supports the latest technology for mission planning in robotics while remaining simple to use.
\section{System Architecture and Design} \label{sec:method}
\begin{figure*}[t]
\centering
\includegraphics[width=0.8\linewidth]{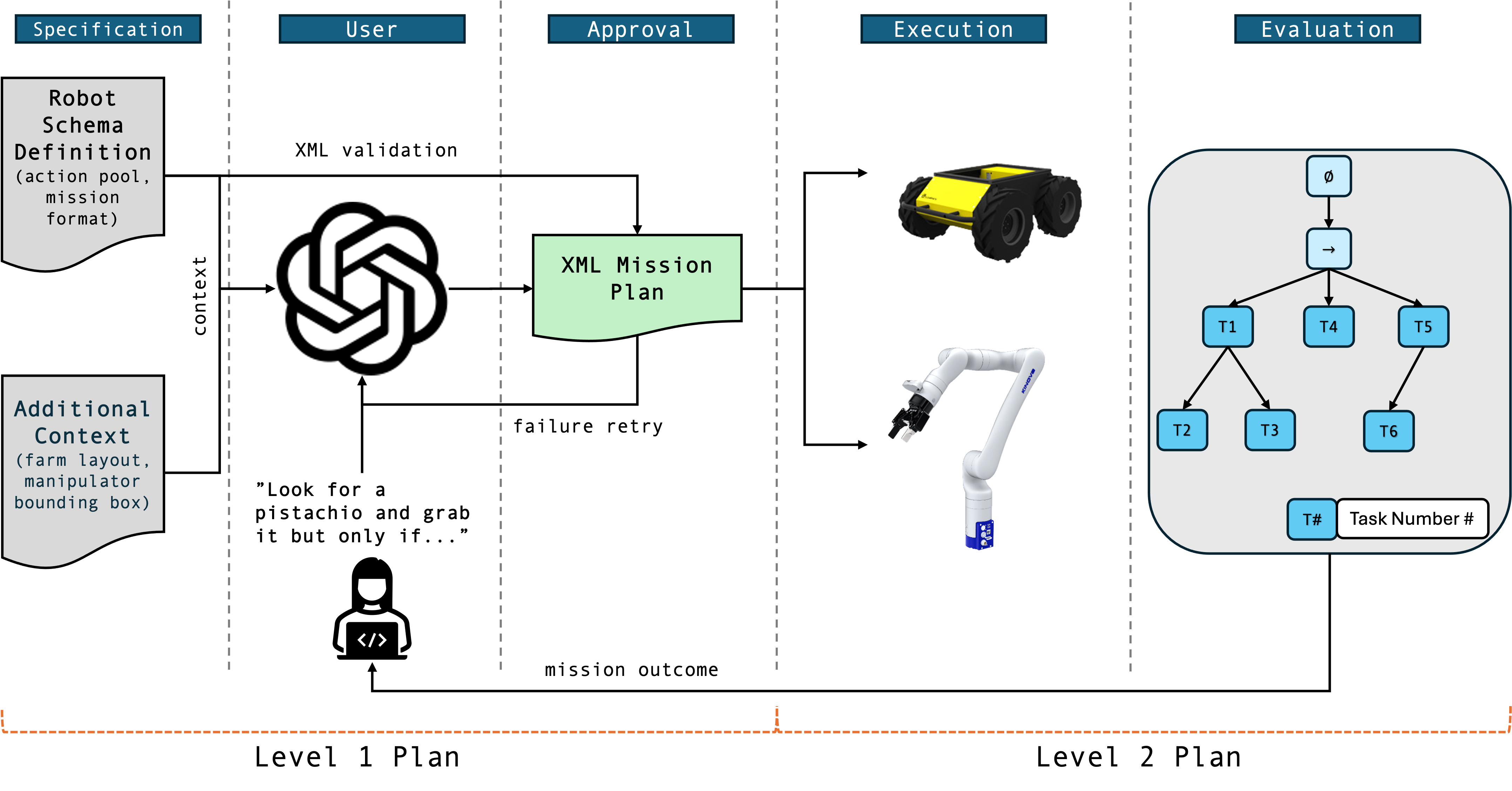}
\caption{Our MP architecture, adapted from \citep{CarpinLLMSubmitted}. This shows how the system interface remains the same with a modular capability during execution.}
\label{fig:gnn-sop-architecture}
\end{figure*}

\subsection{Problem Definition} \label{sec:formal}

Robot mission planning has been defined in multiple ways in the literature. 
We adopt the classic discrete feasible planning formulation (see, e.g., \citep[chap.~2]{LaValleBook}) comprising a non-empty state space $S$, an action space $A$, a transition function\footnote{For brevity, we consider the case where each action $A$ can be applied in each state in $S$, but the formulation can be easily extended to allow state-specific action sets. This extension would introduce the classic concept of \emph{precondition}.} $T:S\times A \rightarrow S$, an initial state $s_0 \in S$, and a set of goal states $S_G \subset S$.
In the feasible planning problem, the objective is to find a sequence of actions that, when applied in order, would transition the system from the initial state $s_0$ to any state in $S_G$. 
Traditionally, one would provide an explicit or implicit representation of $S, A, T, s_0$, and $S_G$, and then use a search algorithm to explore the associated search graph and determine solution existence.
In our system, we formulate the action set $A$ as a set of capabilities coded as Robot Operating System 2 (ROS2) \emph{actions}, each targeting a specific robot platform. 
The key innovation is that once this action pool is adequately represented and provided to the LLM as context, we rely on the LLM to infer the remaining components: $S, s_0, T$, and $S_G$. As elaborated later, this inference draws upon the full context and the user's query. Even more importantly, once the LLM has generated the full context
for the mission planning problem, we also rely on the LLM to determine a plan that is then offloaded to
the robot for execution.

\subsection{One For All Architecture}
We refer the reader to \citep{CarpinLLMSubmitted} for a full discussion of the system architecture of the mission planner, and in this section we briefly summarize it.
Our system consists of a five-stage data pipeline broken up into two task plan subsystems: Level 1 (L1) and Level 2 (L2).
As the NL mission query progresses through the pipeline, it follows the framework provided in the IEEE standard 1872.1-2024 \citep{noauthor_ieee_2024}: specification, user, approval, execution, and evaluation. 
Each of these stages represent a different software module(s) that assists in mission plan decomposition to relevant robot tasks.
Figure \ref{fig:gnn-sop-architecture} sketches the detailed data flow.
The \textit{specification} stage begins the L1 plan decomposition. 
Relevant context files along with the mission plan are requested from the user.
The context files can be anything from world information to robot specifications, but must always contain an XSD file that defines the robot capabilities, i.e., the action set $A$ discussed in Section \ref{sec:formal}.
Not only does the XSD define action set, but also constrains the resultant XML mission plan to take the shape of a behavior tree.
This constraint helps simplify the manner in which L2 plan is decomposed.
The XSD file defined for the Kinova KORTEX was written to match the framework suggested in \citep{noauthor_ieee_2024}, thus enabling new robots to be included up simply following a standardized methodology.
Referred to as atomic actions, these capabilities defined in the XSD are the only code that require update when introducing a new robot with a new set of capabilities, i.e., actions available to the planner.
These updates can include robot action definitions in XSD and their associated parameter definitions among other state information.  
For example, we implemented a low-level primitive for detecting an object that corresponds to a ROS2 node action for detecting an object via the YOLOv11 off-the-shelf network. 
We assigned generic parameters such as object name and color as available parameters to be used, but not required.
These definitions can be expanded per use case and are the only ones requiring manual coding since each XSD tag maps to a ROS2 node action.

After mission prompt and context are sent by the user, this information goes to the LLM. 
In our current implementation
OpenAI's ChatGPT is used to generically parse the NL L1 plan, though any LLM can be easily substituted
and we are currently expanding the system to include additional LLMs such as Anthropic's Claude.
This phase is referred to as the \textit{user} phase.
The output is a L1 XML mission plan compliant with \citep{noauthor_ieee_2024}. 
It is at this stage that the LLM parses the context and the query and infers the remaining components of the planning problem, i.e., $S,T,s_0$ and $S_G$. 
Additionally, the XML returned by the LLM encodes the plan to solve the MP problem inferred from the context and the user query.

Next is a stage called \textit{approval}.
The approval phase is critical in ensuring that whatever mission plan is generated by the LLM will properly convert into a L2 mission plan.
This is done by using an XML validation engine to compare the XSD to the XML, verifying its syntactic correctness.
Should the LLM return an incorrect plan, the \textit{approval} module will request a rewrite with the error log from the validation engine.
Finally, the L1 plan is passed through a TCP socket out to the computer hosting the control system, if not the same.

Last is task \textit{execution} and \textit{evaluation}. 
\textit{Execution} begins with XML conversion into a behavior tree creating our first L2 plan sequence. 
Supporting conditional task execution, the behavior tree encodes the relationship between tasks for task results to be evaluated at run time.
In tandem with the \textit{evaluation} stage, each task is assigned a software module based on the available robot actions defined in the XSD. 
When a task enters the queue, the behavior tree identifies the necessary inputs and begins the task.
Upon completion, the result is evaluated by the behavior tree and the next task is selected.
These two stages are critical in understanding the flexibility and power of the system.
\citep{kambhampati_llms_2024} critically reviews that leaving only an LLM to plan can result in poor performance.
In \citep{CarpinLLMSubmitted} we also cite a similar experience with LLM-only performance.
As such, this architecture allows for optimized software modules to manage task execution and or evaluation.
Therefore, while we stand to gain on the generality of the front-end interface, we do not lose on performance.

\subsection{Key Design Updates}
While the majority of the architecture remains the same as the previous implementation in \citep{CarpinLLMSubmitted}, there are several differences that support a distributed robot environment. 
First, instead of providing context to our LLM planner for a single robot, we provide all relevant files and allow the LLM agent to select which is needed for a given mission. 
This includes passing XSD files for all robots, which include their action pools, virtual farm representations, robot arm constraint definitions, or anything else required. 
In our experiments, however, we limit these files to XSD and Geographical JavaScript Object Notation (GeoJSON) type files only. 
We will demonstrate, in Section \ref{sec:results}, the implications of this design.
The use of XSD for schema definition remains critical in this architecture as it allows the LLM to extract the unspecified components of the planning domain.
We amended the previous XSD to be more concise with minor syntax and structural updates.
The only functional updates made were in the new Kinova KORTEX schema that added it's action pool.
\section{Results} \label{sec:results}

\begin{table*}[t]
\begin{center}
    \begin{tabular}{ccccc}
    \cline{1-5}
     \multicolumn{1}{c}{Mission Queries} &  \multicolumn{1}{c}{Intended Robot} & \multicolumn{1}{c}{Mission Type} & \multicolumn{1}{c}{Number of Tasks} & \multicolumn{1}{c}{Success?}\\\cline{1-5}
     \hline
     "\textit{Find pistachio and take NBV}" & KORTEX & non-spatial & 2 (1 conditionals) & True\\
     "\textit{Find pistachio and pick it}" & KORTEX & non-spatial & 2 (1 conditionals) & True\\
     "\textit{Pistachio NBV conditionals (Figure \ref{fig:behavior_tree})}" & KORTEX & non-spatial & 11 (4 conditionals) & True\\
     "\textit{Picture, temperature, co2, drive, repeat if low readings}" & Husky & non-spatial & 14 (5 conditionals)& False$^!$ \\
     "\textit{Measure all sensors; if-else clause for every reading}" & Husky & non-spatial & 16 (5 conditionals)& True \\
     "\textit{Turn gripper left (relative movement)}" & KORTEX & spatial &1 (0 conditionals)& False \\
      "\textit{Turn left (absolute movement)}" & KORTEX & spatial &1 (0 conditionals)& True \\
     "\textit{Move in a square}" & KORTEX & spatial & 4 (0 conditionals) & True\\
     "\textit{Find object take NBV. If not present, find another.}" & KORTEX & spatial & 3 (2 conditionals)& True$^*$ \\ 
     "\textit{Move in a square and take pictures}" & Husky & spatial & 8 (0 conditionals) & False\\

     \hline
    \end{tabular}
    \caption{
    Shortened queries with intended robot, mission type, complexity, and results. 
    \newline $^*$ see Section \ref{ref:nbv} for more details. $^!$ originally passed, but now fails. 
    \newline More experiments and  prompts can be found at 
    \tt {https://ucmercedrobotics.github.io/one4all.html}
    } \label{table:results}
    \end{center}
\end{table*}

\subsection{Experimental Setup}
On the front end, the system uses OpenAI's ChatGPT GPT-4o-2024-11-20 with a temperature of $0.2$ and max response tokens of $4096$.
On the back end, we have two robots that receive the decomposed task list and act upon the missions.
First is the Clearpath Husky: a 4-wheeled robot equipped with sensors for temperature, thermal vision, and $CO_2$ flux sensor.
Available actions are go to GPS location, read temperature, take thermal image, and measure $CO_2$ flux.
Using Clearpath's control ROS2 framework, we have written ROS2 nodes for movement, localization, and sensor use.
Second is the Kinova KORTEX Gen3 manipulator: a robot arm with 6 DOF and a vision bracelet attached above the gripper.
Available actions are go to position -- relative or absolute position -- detect object, and capture images.
The KORTEX comes packaged with ROS2 nodes for control and vision. Our robot is mounted on an Amiga 
robot by Farm-ng.
While this paper does not explore applying the architecture to an Amiga, the assumption is that if heterogeneous robotic control is proven, this package can be applied to any mobile or manipulator robot.
% \begin{figure}[htb]
%     \centering
%     \includegraphics[width=0.8\linewidth]{figures/20250224_160506.jpg}
%     \caption{Our experimental platform consists of a Gen 3 manipulator by Kinova  mounted on an Amiga
%     mobile base by Farm-ng.}
%     \label{fig:kinova}
% \end{figure}
For control of the arm, we wrote a YOLOv11 ROS2 node for object detection and a node for simplified inverse kinematic control in 3D space using MoveIt2.
MoveIt2 extends these nodes as a kinematic control library for arm movement planning and execution that simplifies KORTEX interactions. 
These nodes extend the L2 plan derived from our LLMs L1 plan.
% We containerized the above implementation in Docker for simplicity.

The only input to the system remains context and mission prompt.
For the context, we only supplied 3 files.
For the Husky, we provide an XSD with task relationships defined along with robot action pool.
We also provide additional context in the form of a virtual farm GeoJSON.
For the Kinova KORTEX, we simply provide the XSD.
These mission prompts aim to explore the strength and spatial awareness of the planner between two possible robots without the need to reconfigure.
Plans are executed only with feedback from task outcomes, no LLM feedback.

\subsection{MP Solutions}
In Table \ref{table:results}, we show the full list of queries demonstrating the flexibility and capability of the architecture.
The number of tasks count includes total atomic tasks and conditionals. 
The intended robot defines which robot should ideally be selected for the mission. 
The success column is defined as semantically representing the prompt as defined by the user.
Note that all missions were manually reviewed for semantic success.
With the missions emulating planning in an orchard, we show how the system is capable of complex mission design.
We demonstrate mission generation for both spatial and non-spatial queries.
However, we will showcase a need for supporting software to overcome issues in spatial planning.

\begin{figure}[htb]
\centering
\includegraphics[width=0.9\linewidth]{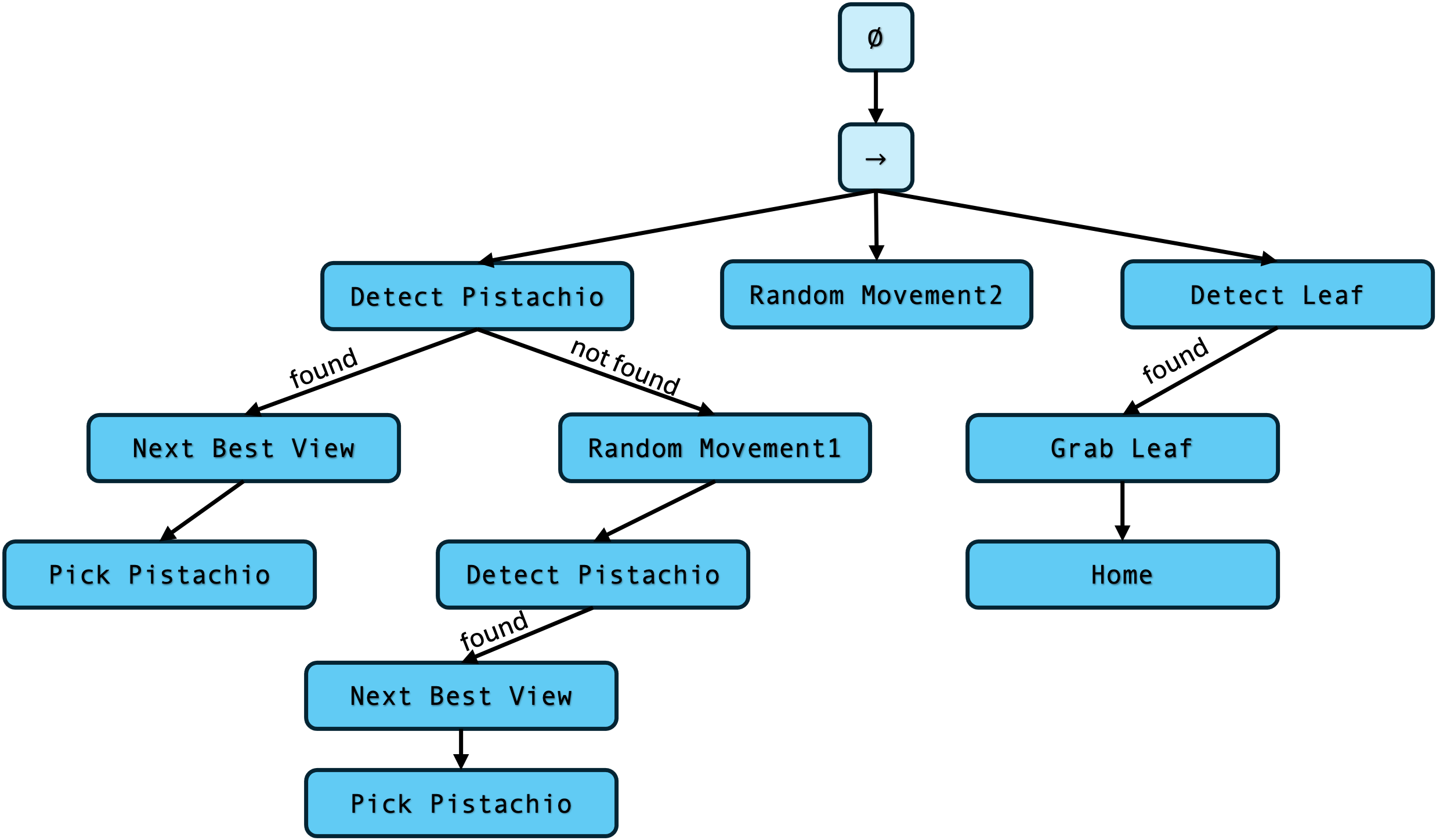}
\caption{Behavior tree visualized from sample query.}
\label{fig:behavior_tree}
\end{figure}

While Table \ref{table:results} only shows a handful of previously examined queries from \citep{CarpinLLMSubmitted}, all experiments were re-run to ensure no regressions occurred.
The system continues to show the ability to generate complex behavior tree plans for both the Husky and the Kinova with limited guidance.
As an example, one of the more complex queries requires multiple conditional actions and does not explicitly describe the behavior tree.
The query, drawn as a behavior tree in Figure \ref{fig:behavior_tree} is as follows, ``\textit{Look for a pistachio. If you find one, take NBV and pick it. If not, make a random move to find another one. If you find this one, NBV and pick. Next, move to another random spot and look for a leaf. If found, grab the leaf and move home.}'' 
Next Best View (NBV), is defined in the action pool as a series of Point Cloud images around an object. 
See Section \ref{ref:nbv} for more details.

It is important to understand that when prompted with a mission, context for all available robots is given to the LLM, yet none of the missions explicitly state which robot should be given the task.
With this information, the system appropriately selects the right robot and designs the mission in accordance with the schema in almost all cases.
For example, if asked, ``\textit{Could you turn around and look for the red cup? If you don't find it, turn back around,}'' the system demonstrates understanding.
In this case, the agent selects Kinova even though the query is ambiguous.
However, there are limits to these types of queries. 
Our results show that should a statement be overly ambiguous, the system responds with no mission plan and tells the user it does not understand.
This problem is remedied by keying the robot explicitly as a part of the query or by being more descriptive. 

\subsection{Atomic Tasking and Modularity} \label{ref:nbv}
As defined by \citep{noauthor_ieee_2024}, a robot task execution system must define atomic tasks. 
Though the primitiveness of these tasks is not explicit.
Instead of allowing the LLM to have access to the most primitive actions in robotics, we abstract away lower-level actions in favor of higher-level ones.
This is demonstrated easiest by the "detect object" action.
While there have been papers \citep{chen_driving_2024, xu_drivegpt4_2024, elhafsi_semantic_2023} that do semantic reasoning on images, there are proven solutions, such as YOLO, that are very simple to use. 
% Not to mention that this application does not have an option for real-time feedback with the LLM.
In the case of asking the LLM to spatially plan using the Kinova at mission initialization, we need to understand the concept of a NBV. 
In literature, NBV \citep{burusa_attention-driven_2024, gao_take_2024} is proven to be a non-trivial task.
We experiment asking the LLM to generate a sequence of camera poses that would best cover an object, for which it was not able to. 
The LLM could generate poses, but struggled to understand the quaternion algebra required to generate 3D Point Cloud reconstructions.
The missions that came out were unusable as, in the most trivial requests, the LLM could not grasp the fact that different joints on a manipulator have different axes orientations -- \textit{even when explicitly described}. 

\begin{figure}[h]
    \centering
    \begin{subfigure}[b]{0.24\textwidth}
        \centering
        \includegraphics[width=\linewidth]{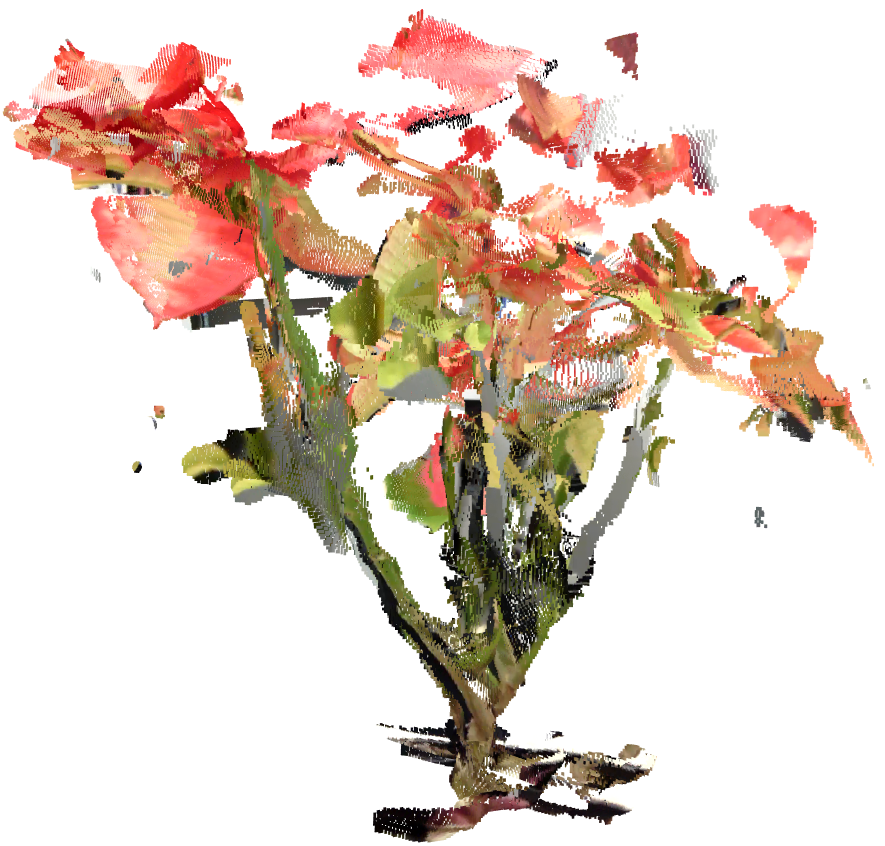}
        \caption{3D Point Cloud of a plant}
        \label{fig:pointcloud}
    \end{subfigure}%
    \hfill
    \begin{subfigure}[b]{0.24\textwidth}
        \centering
        % Specify both width and height to force the desired dimensions
        \includegraphics[width=\linewidth, height=4.15cm]{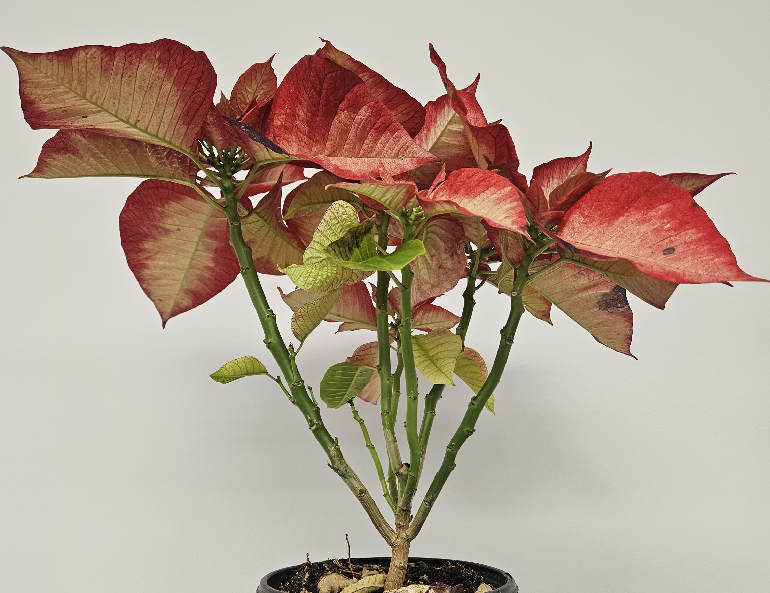}
        \caption{Color image of a plant}
        \label{fig:rgbimage}
    \end{subfigure}
    \caption{On the left is a NBV 3D Point Cloud reconstruction of the image on the right}
    \label{fig:nbv}
\end{figure}

Instead of trying to explain NBV to the LLM, we experiment with defining an atomic task to represent NBV.
This abstracts away positional and rotational orientation from the LLM and passes it to a module more equipped to handle such complexity.
In doing so, the architecture remains general enough for a nontechnical user to create missions.
The modularity in both the L1, XSD action pool, and L2, ROS2 node, phases is the key contribution of this architecture. 
In Figure \ref{fig:nbv}, we see the result of an experiment that uses a plug-and-play module for NBV combined with a higher level action in the pool for the LLM to select NBV.
Figure \ref{fig:nbv} shows the reconstruction of a potted
plant in an indoor environment.
By decoupling L1 and L2 tasking, this enables end users to only be required to know about general capabilities of their robot and supporting software.
Then, those responsible for driver and functional support can update modules without fear of breaking the control pipeline.

% \begin{figure}[h]
%         \centering
%         \includegraphics[height=5cm]{figures/treescan.png}   
%     \caption{Reconstruction of a tree in an outdoor environment.}
%    \label{fig:tree}
% \end{figure}

\subsection{Limitations} \label{sec:limits}
% Originally, we began the experimentation using OpenAI's newest LLM, o1-mini-2024-09-12, but quickly realized that it was not capable of efficiently generating missions. 
% While after several iterations of error feedback with the LLM o1-mini would eventually generate a valid mission, this process took an average of four tries.
% All of the missions generated the same with the exception of how many tries it took the system to reach a valid mission.
% As such, we opted for GPT-4o.

For queries with vague language, generating the expected mission was difficult for the LLM.
For example, a query specified the identification of an orange tree, which was not included in the virtual farm context file, even though the prompt used available actions.
It seems when plans are requested that relate somewhat to the context -- dealing with fruit with a virtual farm filled with pistachios -- the LLM has difficulty realizing that the robots are often capable of actions even without context.
As in an exploratory task where information is unknown but plausible: finding an orange tree in a pistachio farm.
However, when asked to interact with completely adjacent objects, the system does not seem to struggle.
This is demonstrated by successful missions generated asking for inanimate objects such as cups or plants that are not included in the virtual farm file.
The confusion was overcome simply by expressing more explicit requirements into the mission query, such as to generate a mission regardless of context. 

Asking the system to understand kinematics proved to be the only problem that could not be overcome simply. 
As in the route optimization problem in \citep{CarpinLLMSubmitted}, LLMs seemingly do not understand what types of motion planning exist in robotics.
Even for a prompt as simple as asking the planner to turn the gripper left proved difficult. 
We have clearly shown that spatial reasoning, even in the simplest forms, is quite difficult for our LLM-backed agent.
% However, the strength of the architecture doesn't lie in the LLM, but rather the ability to enhance any mission with considerably stronger software.
We overcame this in \citep{CarpinLLMSubmitted} by adding a graph neural network (GNN) to the \textit{evaluation} stage of the pipeline for solving graph based problems on the Husky. 
% The results of that experiment using this update system are shown in Figure \ref{fig:path}.
We again lean on the modularity and flexibility of this architecture to allow for varying levels of task support.
See Section \ref{ref:nbv} for details on kinematic support. 
\section{Conclusions and Future Work} \label{sec:conclusions}
In this paper we presented a MP design that is flexible enough to support different robots, but powerful enough to handle real life applications. 
The solution we proposed not only enables non-specialists
to plan robot missions but doing so for multiple types of robots from a single interface.
Our experimentation has shown that this architecture handles complex missions but has its limitations when dealing with spatial awareness. 
We overcame many of the problems with more explicit mission prompting and more capable supporting software, but gaps still remain to make the system seamless. 
Additionally, vagueness in queries lead to the system generating a valid mission but one that is slightly different from what the user had intended. 
As we have combined a set of robots into a single mission planning system, this begs the question of being able to run a single query to generate missions for multiple robots. 
Often times, robot fleets are intended to cooperate, but these experiments have only shown individual behavior.
Future research will focus on the vagueness of missions and how to close the loop on uncertain plans.
Also, we plan to experiment with decoupling single mission prompts into one or more mission plans for a fleet of robots.

\bibliography{report}

\begin{thebibliography}{21}
\providecommand{\natexlab}[1]{#1}
\providecommand{\url}[1]{\texttt{#1}}
\providecommand{\urlprefix}{URL }
\expandafter\ifx\csname urlstyle\endcsname\relax
  \providecommand{\doi}[1]{doi:\discretionary{}{}{}#1}\else
  \providecommand{\doi}{doi:\discretionary{}{}{}\begingroup \urlstyle{rm}\Url}\fi

\bibitem[{Ahn et~al.(2022)}]{ahn_as_2022}
Ahn, M. et~al. (2022).
\newblock Do {As} {I} {Can}, {Not} {As} {I} {Say}: {Grounding} {Language} in {Robotic} {Affordances}.
\newblock ArXiv:2204.01691 [cs].

\bibitem[{Burusa et~al.(2024)}]{burusa_attention-driven_2024}
Burusa, A.K. et~al. (2024).
\newblock Attention-driven next-best-view planning for efficient reconstruction of plants and targeted plant parts.
\newblock \emph{Biosystems Engineering}, 246, 248--262.

\bibitem[{Chen et~al.(2024)}]{chen_driving_2024}
Chen, L. et~al. (2024).
\newblock Driving with {LLMs}: {Fusing} {Object}-{Level} {Vector} {Modality} for {Explainable} {Autonomous} {Driving}.
\newblock In \emph{Proceedings of the {IEEE} {International} {Conference} on {Robotics} and {Automation}}, 14093--14100.

\bibitem[{Elhafsi et~al.(2023)}]{elhafsi_semantic_2023}
Elhafsi, A. et~al. (2023).
\newblock Semantic {Anomaly} {Detection} with {Large} {Language} {Models}.
\newblock ArXiv:2305.11307 [cs].

\bibitem[{Emsley(2023)}]{emsley_chatgpt_2023}
Emsley, R. (2023).
\newblock {ChatGPT}: these are not hallucinations – they’re fabrications and falsifications.
\newblock \emph{Schizophrenia}, 9(1), 1--2.
\newblock Publisher: Nature Publishing Group.

\bibitem[{Gao et~al.(2024)}]{gao_take_2024}
Gao, S. et~al. (2024).
\newblock Take {Your} {Best} {Shot}: {Sampling}-{Based} {Next}-{Best}-{View} {Planning} for {Autonomous} {Photography} \& {Inspection}.
\newblock ArXiv:2403.05477 version: 1.

\bibitem[{Huang et~al.(2022)}]{huang_language_2022}
Huang, W. et~al. (2022).
\newblock Language {Models} as {Zero}-{Shot} {Planners}: {Extracting} {Actionable} {Knowledge} for {Embodied} {Agents}.
\newblock ArXiv:2201.07207 [cs].

\bibitem[{IEEE(2024)}]{noauthor_ieee_2024}
IEEE (2024).
\newblock {IEEE} {Standard} for {Robot} {Task} {Representation}.
\newblock \emph{IEEE Std 1872.1-2024}, 1--32.

\bibitem[{Janßen et~al.(2023)}]{jansen_can_2023}
Janßen, C. et~al. (2023).
\newblock Can {ChatGPT} support software verification?
\newblock ArXiv:2311.02433 [cs].

\bibitem[{Jousselme et~al.(2023)}]{jousselme_uncertain_2023}
Jousselme, A.L. et~al. (2023).
\newblock Uncertain about {ChatGPT}: enabling the uncertainty evaluation of large language models.
\newblock In \emph{2023 26th {International} {Conference} on {Information} {Fusion} ({FUSION})}, 1--8.

\bibitem[{Kambhampati et~al.(2024)}]{kambhampati_llms_2024}
Kambhampati, S. et~al. (2024).
\newblock {LLMs} {Can}'t {Plan}, {But} {Can} {Help} {Planning} in {LLM}-{Modulo} {Frameworks}.
\newblock ArXiv:2402.01817 [cs] version: 2.

\bibitem[{Kannan et~al.(2024)}]{kannan_smart-llm_2024}
Kannan, S. et~al. (2024).
\newblock {SMART}-{LLM}: {Smart} {Multi}-{Agent} {Robot} {Task} {Planning} using {Large} {Language} {Models}.
\newblock ArXiv:2309.10062 [cs].

\bibitem[{LaValle(2006)}]{LaValleBook}
LaValle, S. (2006).
\newblock \emph{Planning algorithms}.
\newblock Cambridge academic press.

\bibitem[{Liang et~al.(2023)}]{liang_code_2023}
Liang, J. et~al. (2023).
\newblock Code as {Policies}: {Language} {Model} {Programs} for {Embodied} {Control}.
\newblock In \emph{Proceedings of the {IEEE} {International} {Conference} on {Robotics} and {Automation}}, 9493--9500.

\bibitem[{Mower et~al.(2024)}]{mower_ros-llm_2024}
Mower, C. et~al. (2024).
\newblock {ROS}-{LLM}: {A} {ROS} framework for embodied {AI} with task feedback and structured reasoning.
\newblock ArXiv:2406.19741 [cs].

\bibitem[{Pelucchi and Valdenegro-Toro(2023)}]{pelucchi_chatgpt_2023}
Pelucchi, M. and Valdenegro-Toro, M. (2023).
\newblock {ChatGPT} {Prompting} {Cannot} {Estimate} {Predictive} {Uncertainty} in {High}-{Resource} {Languages}.
\newblock ArXiv:2311.06427.

\bibitem[{Purcell and Neubauer(2023)}]{PURCELL2023100094}
Purcell, W. and Neubauer, T. (2023).
\newblock Digital twins in agriculture: A state-of-the-art review.
\newblock \emph{Smart Agricultural Technology}, 3, 100094.

\bibitem[{Ray(2023)}]{ray_chatgpt_2023}
Ray, P. (2023).
\newblock {ChatGPT}: {A} comprehensive review on background, applications, key challenges, bias, ethics, limitations and future scope.
\newblock \emph{Internet of Things and Cyber-Physical Systems}, 3, 121--154.

\bibitem[{Sani et~al.(2024)Sani, Sgorbissa, and Carpin}]{CARPINICRA2024D}
Sani, E., Sgorbissa, A., and Carpin, S. (2024).
\newblock Improving the ros 2 navigation stack with real-time local costmap updates for agricultural applications.
\newblock In \emph{Proceedings of the {IEEE} International Conference on Robotics and Automation}, 17701--17707.

\bibitem[{Xu et~al.(2024)}]{xu_drivegpt4_2024}
Xu, Z. et~al. (2024).
\newblock {DriveGPT4}: {Interpretable} {End}-to-end {Autonomous} {Driving} via {Large} {Language} {Model}.
\newblock ArXiv:2310.01412 [cs] version: 4.

\bibitem[{Zuzu\'{a}rregui and Carpin(2025)}]{CarpinLLMSubmitted}
Zuzu\'{a}rregui, M.A. and Carpin, S. (2025).
\newblock Leveraging {LLM}s for mission planning in precision agriculture.
\newblock In \emph{Proceedings of the {IEEE} International Conference on Robotics and Automation}, 7146--7152.

\end{thebibliography}

\end{document}